# Beyond Black Boxes: An Ontology Based Vision Framework for Transparent Fruit Disease Detection

M.J. Marapperuma
*Department of Computer Engineering*
*General Sir John Kotelawala Defence University*
Colombo, Sri Lanka
39-bce-0006@kdu.ac.lk

B.V.K.I. Vidanage
*Department of Computer Science*
*General Sir John Kotelawala Defence University*
Colombo, Sri Lanka
vidanage_bvki@kdu.ac.lk

***Abstract*—Efficiency and reliability in the timely identification of phytopathogens would help minimize crop loss in agricultural sectors but existing deep learning based methods, although accurate, would require a large annotated dataset and are black-boxed in their operation, thus, restricting flexibility and transparency. The current paper presents a hybrid detection framework that is able to overcome these drawbacks through the unified real-time vision with a semantic knowledge base. Web Ontology Language (OWL) was used to develop a comprehensive ontology of the disease of fruits that formally codifies phytopathological knowledge, such as symptom phenotype, spectral phenotype (HSV), and morphological phenotype. Such knowledge-guided configuration guides the image processing pipeline, making it possible to diagnose using this knowledge under expert specified rules without a large training dataset. The ontology underwent a computationally verified test of its structural soundness and commonsensical practical usefulness and scored an overall of 7.4/10. On the public set of MangoFruitDDS, the diagnostic performance was an F1 Score of 85.00%, precision of 85.00%, and a recall of 85.23%. These findings are very competitive and similar to other known deep learning models like the VGG16 that prove that the knowledge-driven approach provides an accurate, transparent, and scalable solution on automated fruit disease diagnosis.**

## I. INTRODUCTION

Accurate and early detection of plant diseases during fruit production are part of the whole process of protecting food security and reducing the losses incurred by the farming industry. The manual inspection that most of us are accustomed to is highly labour-intensive techniques that are inherently error-prone thus an automatic computer vision system was being stimulated. This conversion in the contemporary literature describes this progression in the way that classical image-processing routines that have been demonstrated to be successful at certain tasks are now being converted into data-driven and deep learning paradigms [1], [2], [6]. Convolutional Neural Networks (CNNs) have become the best practice and currently achieved high accuracies on mangoes, guavas, and citrus disease identification [2], [7].

However, these deep learning models require large bodies of well-labelled image data in order to be effective, which hinders their implementation due to the cost of collecting such data, training expenses and the resultant black box nature which faults the diagnostic process [3], [4], [5], [8]. In addition, these models often show poor ability to generalize to new diseases or small changes in symptoms without years of re-learning. This limitation presents a critical concern to have an intelligent system that would precede a combination of domain experience, explanations, and flexibility, and at the same time reduce the dependence on the large training sets.

A hybrid architecture, which integrates computer vision and semantic reasoning, will be sufficient to address this need. Using a formal ontology, authoritative phytopathological knowledge is encoded into our system, thus serving to instruct real time detection and classification processes. The framework supports a knowledge-based, transparent process of diagnosis by formalizing diseases, related symptoms, and ways of visual representing, i.e. colour scope of HSV and morphology of spots, in a knowledge base. This paper explains the make-up of this hybrid system, outlines its ontological premises, and specifies a quantitative measure of the ontological quality that can be used to show that it is fit to be applied in practical and automated classification of fruit disease.

## II. LITERATURE REVIEW

Automated disease detection on fruits has gained a wide interest with both the conventional image processing and recently developed deep learning techniques. Initial studies were based on hand crafted features and simple classifiers. As another example, morphological filtering, color binarization, and global thresholding were applied to mango images with grading accuracies of nearly 97%. On the same note, regression based monochrome (grayscale) vision has been used with an accuracy of 97.9% on defect detection in mangoes [7]. Features of color, texture, shape have been chosen by an optimization approach such as artificial - bee - colony and discriminated by Support Vector Machines (SVMs) to express more complex pipelines in mango skin diseases. It was recently stated in a survey that in the hyperspectral domain even, Convolutional Neural Networks (CNNs) are capable of classifying several conditions of the citrus peel (canker, spots, etc.) with an almost near-perfect precision of 99.79%, displaying the strength of deep features over simple ones. Overall, the research on fruit diseases presents a clear tendency where traditional methods which are based either on thresholds or texture analysis can be quite good at example specific tasks, but nowadays, end-to-end CNNs and transfer learning take center-stage owing to a better of accuracy in diverse situations [1], [2].

In the case of mango, late 2010s saw frequent practices of hand-crafting approaches. Kent mangoes using merely simple geometry and color-threshold processing obtained 97% correct classification. In another, 97.9% was reported using a monochrome imaging system with multivariate regression. In comparison, mango is addressed today with deep CNN. A 13 layers CNN was tested using 100 images of Kent mangoes (50 healthy, 50 with surface defects). The number of images was increased five times via data augmentation (rotations, flips, etc.) and totalled 800. The CNN was able to attain nearly 98% precision in differentiating the good fruit and defective fruit [6], [7]. More importantly, the deep model did not involve any manual segmentation or feature engineering, but the authors observed that the small size of datasets required excessive augmentation, which made it too expensive to train. On the whole, the mango literature represents the transition of classical vision pipelines to CNNs to detect the defects in high accuracy.

CNN based techniques have also been used to target guava fruits. In one study, a local data of images of diseased guava fruit was collected and data augmentation was performed followed by classification. Seven common architectures (AlexNet, SqueezeNet, GoogLeNet, ResNet-50, ResNet-101) were evaluated and the best accuracy rate (97.7%) was obtained using ResNet-101 [7], [9]. The images were preprocessed in this research (histogram equalization, unsharp masking) and disease classes (canker, rust, etc.) were used as the output labels. The accuracy is very high, which implies that deep features are very applicable to guava diseases. Previously, there was limited literature about more classical methods regarding guava; one of them used basic color-transformation segmentation and SVM/k-Nearest Neighbours (kNN) classification, albeit which provided significantly lower results [2]. Clearly, today, deep CNN and transfer learning are favouring guava.

Passion fruit has not been given a lot of attention until recently. The object-detection model was trained on a large labeled dataset (of the order of 10,000 images) of passion fruits of five different disease types. The authors developed a light - weight CNN that offers a novel mechanism of sparse-parallel attention (SPAM) in order to attend multi-scale lesion areas. The SPAM model had very high detector performance (precision 0.93%, recall 0.88% and overall accuracy 0.91%) on test data set [2]. These numbers (F1 score 90%, mAP50 90%) were much higher than the figures of the baseline detectors (e.g., Faster R-CNN, YOLO, SSD) on the same data. This paper is an example of the use of modern deep detectors capable of attention to localize small lesions on fruit when incorporated in complicated scenes. Previous passion - fruit experiments have also employed YOLO - based networks with mAPs between 80% and 83% after special modifications, but the SPAM technique was much more robust on the variable backgrounds.

These trends are confirmed by other fruit studies. Recently, a CNN architecture, with ResNet the core part, was used to analyze 505 apple fruits to categorize the apples into a few diseases/defects, and the accuracy rate was 78.8%. On real citrus fruits, an inception modules-based combination with EfficientNetV2 as an identification backbone got an accuracy of more than 95% in recognizing the disease [1], [9], [10]. Interestingly, in this research, a VGG-U-Net was also utilized to segment the lesions (accuracy of 87.7%) and specifically transfer learning has been used to speed up the network training process. Previous studies applied hyperspectral imaging and a VGG-16 CNN to predict eight prevalent grapefruit peel situations and the model had 99.8% of eventual accuracy yet hyperspectral arrangements are not as effective in the field [2]. This demonstrate that in terms of methods applied on the fruit (apple, citrus, etc.), the deep CNNs initialized with the pre-trained models always outperform the classical ones.

This shows that there was a marked development in its approach. First attempts featuring classical image-processing methods combined with SVM and ML models produced decent results in a restricted set of data, but contemporary efforts have developed a noticeable and, apparently, constant tendency toward deep learning. The current effort incorporates CNNs and transfer learning models, including ResNet, EfficientNet, and the diverse YOLO implementations to achieve the best accuracies as reported on mango, guava, passion fruit as well as other crops. Such performance is acquired at the cost of intensive data collection: majority of these studies use data augmentation or heavily annotated datasets of bursty size to train deep models [1], [2], [3], [4], [8]. However, deep architecture eliminates manual features as well as labour-intensive manual labeling and ad-hoc feature engineering and is appropriate to process multifaceted real-life imagery. Unsolved issues remain, such as access to data consisting of heterogeneous labeled datasets of fruits and subtle early-stage fruit disease symptoms, but existing evidence reveals that the use of advanced CNN architectures is at the current state-of-the-art in terms of fruit disease detection.

## III. PROPOSED SYSTEM AND METHODOLOGY

A hybrid model which uses both computer vision and semantic reasoning has been suggested to use in real-time detection of fruit and disease. The system architecture follows three guiding principles, namely scalability that is supposedly realized by the live-streaming functions, semantic enrichment realized by the ontological knowledge and adaptive processing that can support heterogeneous aspects of computational effectiveness such as constraints. It is hierarchical in structure where various stages are present as shown in Figure 1.

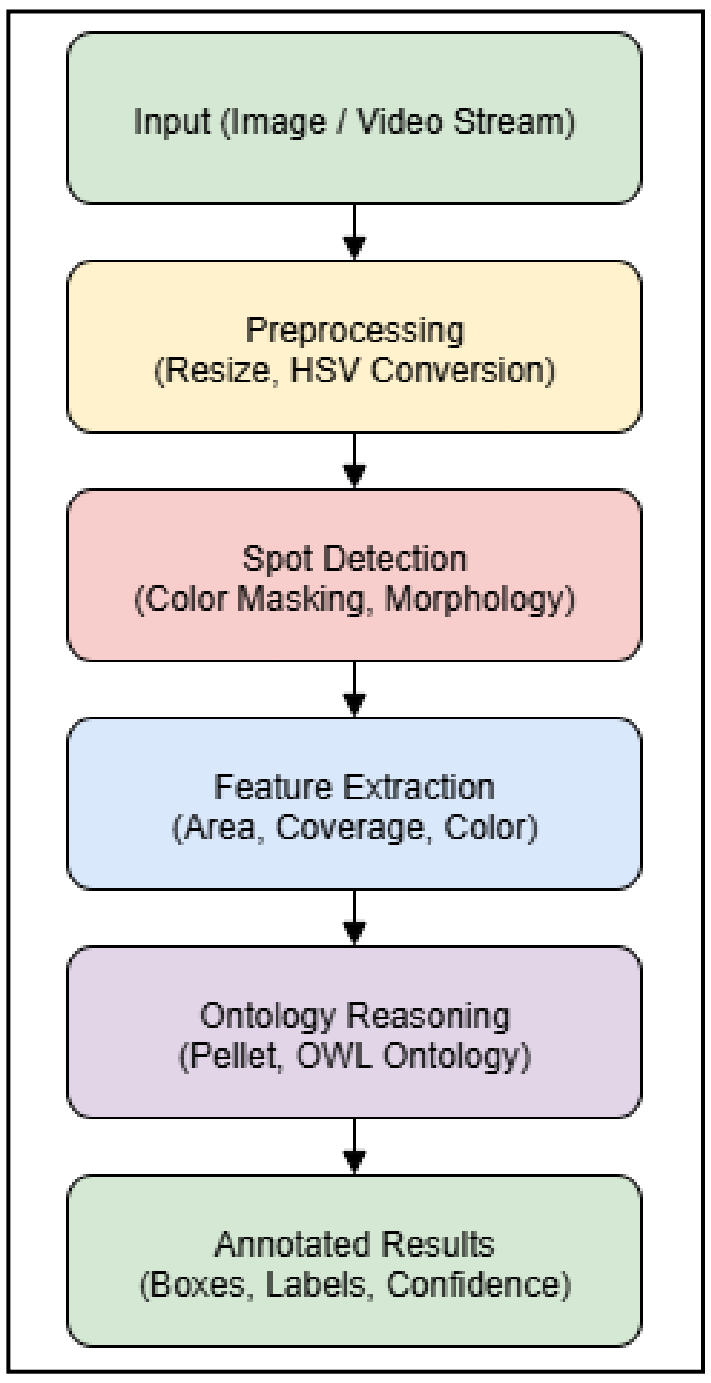

Fig. 1. Overall Framework Design.

The system architecture is based on FruitDiseaseDetector class which plays the role of the main coordinator of computer vision algorithms and the semantic reasoning engine. This integration is encouraged by a bidirectional Vision-Semantic Interface, according to which the visual features can be semantically annotated based on definitive ontological knowledge, but at the same time the vision algorithms can be guided by semantic constraints of the knowledge base. This synergy facilitates a Knowledge-Driven Processing model: graphic attributes particular to each disease, including e.g. HSV color ranges and spot morphology, are seen as a dynamic database to parameterize vision algorithm. This sort of solution ensures that domain expertise keeps detection criteria in place and are available to update without the need to change underlying source code. The sequence of the data flow is through a sequential pipeline with initial steps being acquiring and preprocessing the video frames or the static images. Segmentation on the color basis is then used to determine the possible disease areas and HSV ranges are extracted in real-time based on the ontology itself. The defined features are assigned to ontological concepts to semantically mark them. Pellet reasoner classifies and infers afterward. Aggregated confidence scores are obtained by weighting the results, and informative diagnostic metadata are visualized in the form of display metadata. In a bid to have a robust architecture, fault-tolerant features are used such as a mechanism of switching to using a rule-based classification model in cases where the ontology cannot be found and varying frame rate and resolution according to computational demand.

### *A. Ontology Design and Knowledge Representation*

The knowledge base underlying the currently available system is a part of the ontological infrastructure developed in Web Ontology Language (OWL) to entail the explicit representational coverage of the agricultural field. It uses a hierarchical taxonomy to define the structured construct of the relationship that exist among principal entities namely fruits, diseases, symptoms and pathogens. As part of the classification scheme, Fruit is the host organism, Disease a pathological state categorized by etiology (e.g., FungalDisease), Symptom the manifestations of a pathological state that are observable (e.g., ColorSymptom) and Pathogen the causative agent. Semantic cohesion can actually be achieved by making use of object properties, the hasSymptom, and the causedBy. In order to make the model computation-friendly, it is supplemented with such data properties like hasColorRange (in HSV values), minCoveragePercentage, and spotDiameter which represent measurable properties and thus establish links between the semantic framework and the computer-vision mechanism.

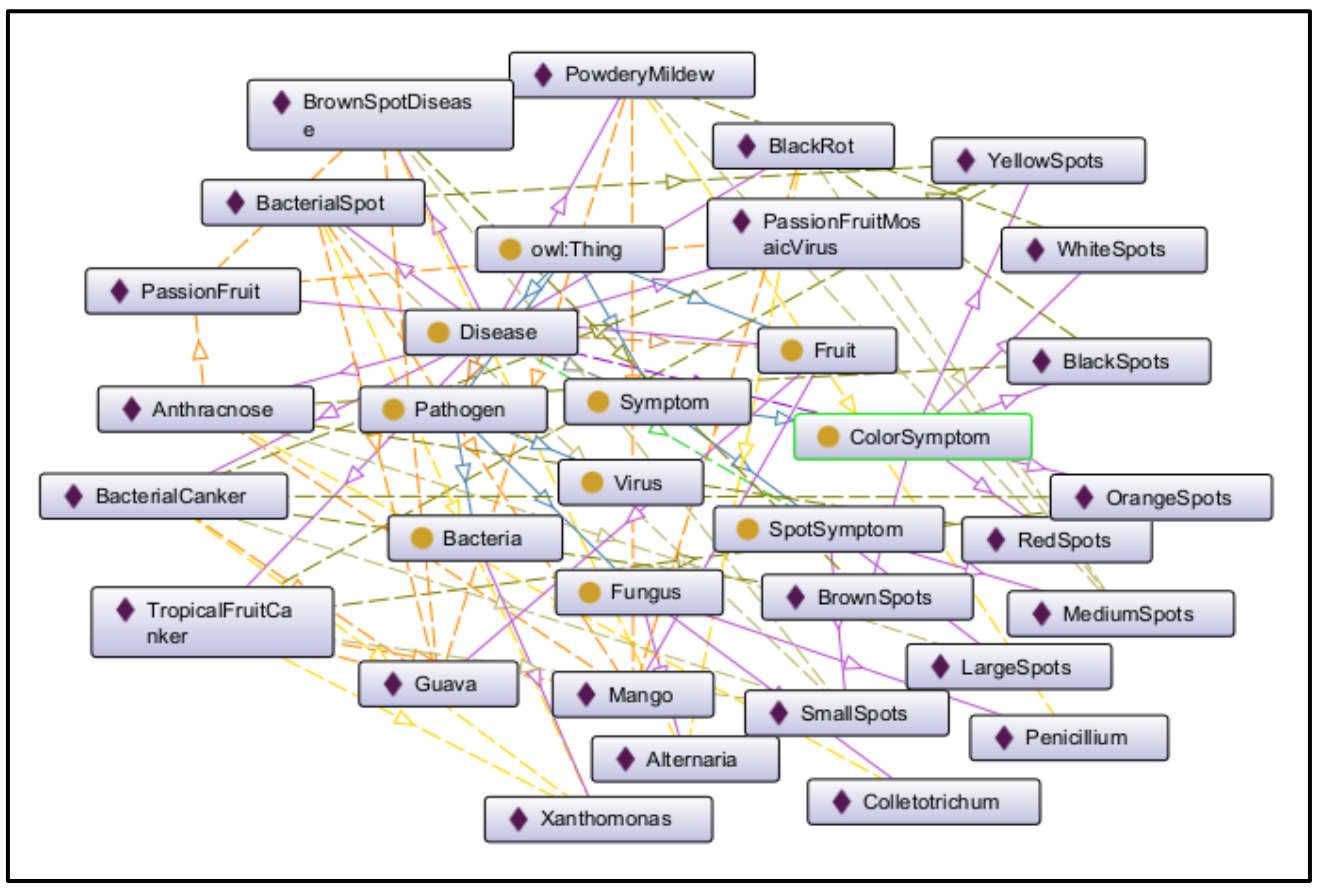

Fig. 2. Ontology knowledge relationship graph developed through OWL.

The curated knowledge base, forming the basis of the current methodology, consists of disease - symptom relationships obtained based on and then tested against the empirical phytopathological literature. More importantly, this resource allows quantitatively parameterizing the visual symptoms to be used in the design of fully automated detection systems. As an example, an entry of Anthracnose is described as associated with the phenotype of brown spot and having Hue-Saturation-Value (HSV) of 8-25 and a minimum of 15% lesion cover. Such visual descriptors are systematically incorporated in the ontology and include specific HSV boundaries of each color segment, morphological constraints on the geometry of symptoms (e.g., diameter/extent of spots, aspect ratio, etc.), and narrative scales associating the expression of such symptoms with progression and severity of illnesses.

A diagnostic system that is able to go beyond a mere rule-based classification was created by combining Pellet, an ontological inference engine that can be automatically used to reason. Pellet is based on subsumption logic and hence the phytopathological conditions are classified in a systematic way seeing complexities in symptomatological combinations. The overall diagnostic conclusion is based on multi-criteria decision process using a weighted confidence scoring algorithm that compresses the evidence of many areas. A composite diagnostic score is computed using the aggregate only weighted inputs with evidence significance relative to each of the factors, morphological compatibility which examines the congruence between the observed lesion geometry and the provided ontology specifications (40%), colorimetric symptom matching, characterized by the correlation of the observed Hue-Saturation-Value (HSV) data with pre-determined range data of color (35%), host specificity by confirming compatibility between the fruit type and published ranges of disease hosts (20%) and influential contextual variables, such as environmental and seasonal variables (5%). The most important aspect of the architecture is that it is designed to support dynamic knowledge evolution, which allows the architecture to be both scalable and maintainable because new disease modules may be easily integrated without making any changes to the underlying structure of the core ontology.

### *B. Vision and Feature Extraction*

The vision pipeline multi-scale processing architecture changes the image resolution, 1080p, 720p, and 480p, on a dynamic basis to optimise the balance of performance versus accuracy to perform processing. Key phases in the preprocessing step consist of colour space transformation of RGB into HSV. The operation gives stronger segmentation using colour and reduces the effect of the lighting variations. A call to morphological operations is then made to improve image quality. Artifacts are suppressed with the kernel-based noise reduction, the boundaries at spots are sharpened applying the gradient-based operators, and the structural gaps in fragmented areas of diseases are filled through the closing operations.

The main component of the algorithm is core detection based on the HSV range-based segmentation using disease-specific ontology. As an illustration, brown spots that are diagnostically relevant to fungal infections will be identified based on a hue range of 8-25, and the black spots that are related to rot are identified in a range of values of 0-45. The result of the segmentation is then fed to improved contour analysis. At this step, area-based filters are used to eliminate noise, geometrical validation is performed with aspect ratio, circularity, and convexity, and hierarchical contour processing is performed in order to identify complicated morphology features, such as concentric rings. In order to facilitate ontological reasoning, the system measures many features encompassing the percentage of disease coverage, the spatial layout of spots as well as morphological features such as the axes length and perimeter-to-area ratios.

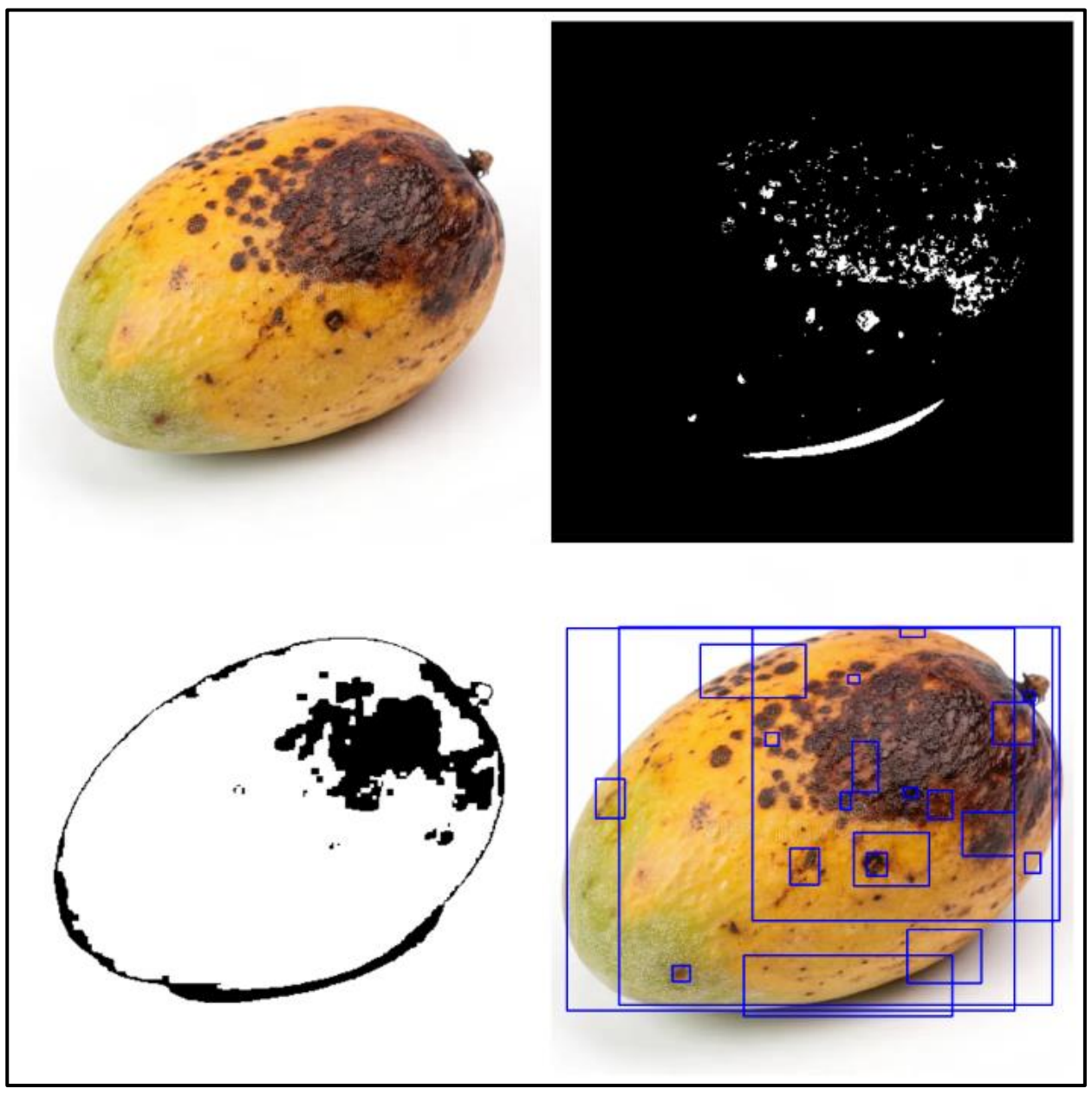

Fig. 3. Step by step feature extraction & disease region detection pipeline.

Adaptive frame processing maintains real-time performance: changing frame processing modes on the basis of performance statistics (frame rate (FPS) and CPU usage), the system sets different frame processing modes according to factors such as frame rate (FPS) and system overhead utilization. Temporal consistency is maximized, so by tracking detected spots across frame, computing unnecessary redundant takes place. As a way of controlling the computational requirements, only a specified number of spots can be detected per frame, while effective memory management strategies are adopted. The architecture can be scaled, multi-threading has been implemented to execute concurrently and is able to live up to GPU-accelerated OpenCV processes.

### C. *Ontology-Driven Disease Reasoning*

These occur by utilizing a reasoning system including a multi criteria scoring algorithm and the amassing of weighted evidence to gain resolute disease classifications. There is a final confidence score that represents each classification and is dependent on a multi-dimensional scoring based on weights that are represented in a four-dimensional matrix that is based on color alignment (35%), morphological compatibility (40%), host-pathogen specificity (20%), and environmental context (5%). Fuzzy logic is incorporated and instead of binary matching, the graded membership levels are defined with trapezoidal membership functions thus allowing fine-grained judgment. This protocol uses a severity based confidence scaling to identify whether various diseases are in their initial stages and tolerance to colour and morphology as natural variations exist.

Pellet reasoner, which is the core of inference, provides full OWL-DL reasoner. It has subsumption reasoning capability which automatically assigns symptoms observed to most precise types of diseases, and it does consistency checking to all inferences generated so as to keep them logically correct.

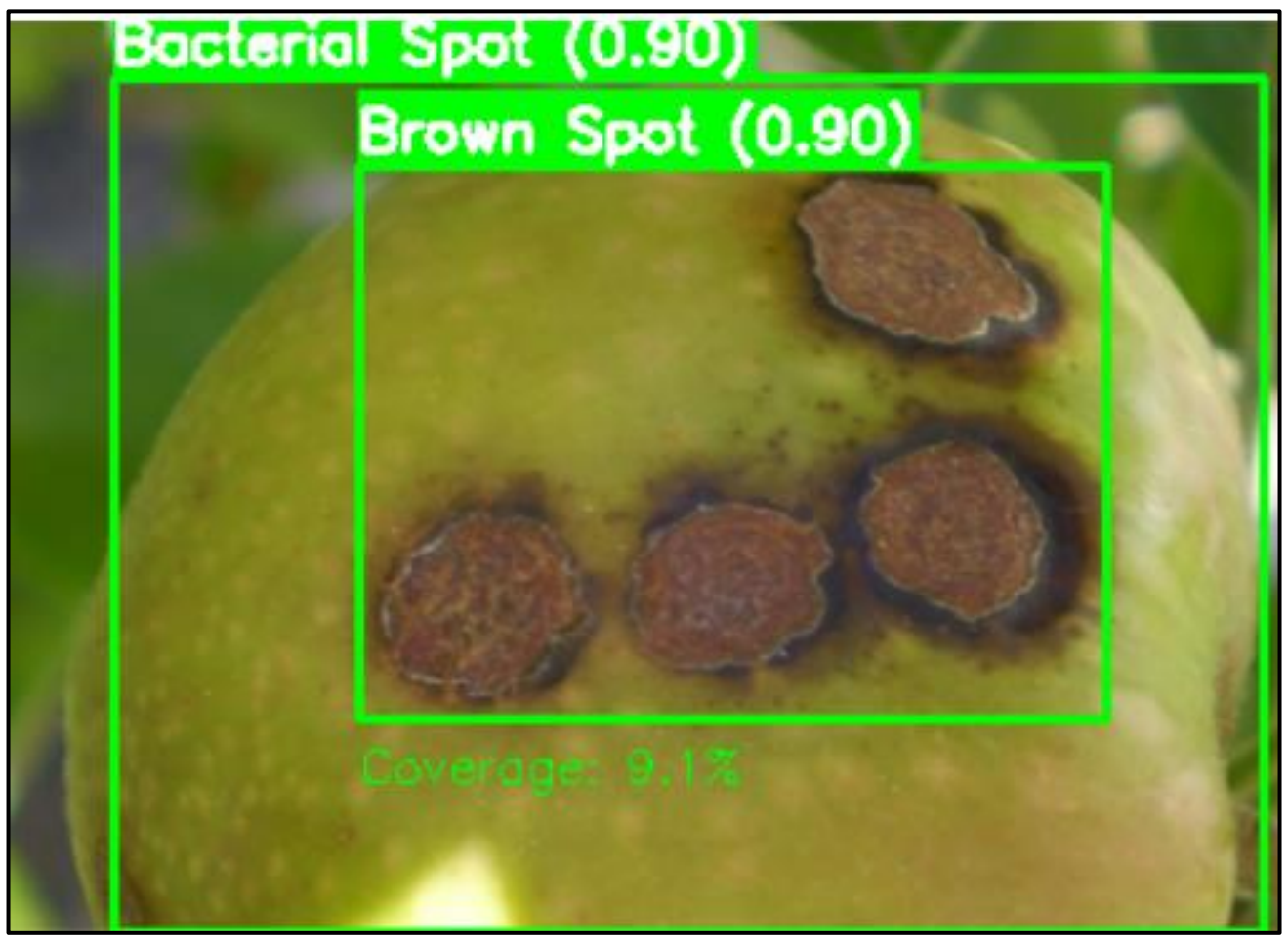

Fig. 4. Final Disease detection result after disease detection & reasoning.

The current system in the classification of disease is designed in several stages that are hierarchically implemented and are well optimised in terms of robustness and efficiency. At its first stage, the pipeline includes the generation of hypotheses, as the visually detectable symptomatology is analyzed with the goal of generating initial disease candidates. Evidence aggregation is in turn applied to this initial list, and is a stage by which all relevant morphological, colorimetric and contextual evidence is accrued and normalised to provide an evidence profile per candidate. In order to ensure diagnostic validity, the candidates are then subjected to semantic validation, which is an important step that entails cross-referencing tests them to a domain-specific ontology in order to test both biological validity and etiological accuracy. The final stage is to employ a confidence-ranking algorithm to rank the accumulated evidence in order to rank all of the validated candidates and derive a final and prioritized list of suspected diagnoses. The system also goes further and makes inferences on practical interventions by checking the ontology on available treatment guidelines as well as preventive strategies of a confirmed disease and its causative agent. All the stages in the piping have been analytically designed to run in real-time by the adoption of the stepwise processing and parallel processing methods.

### D. *Overall System Workflow*

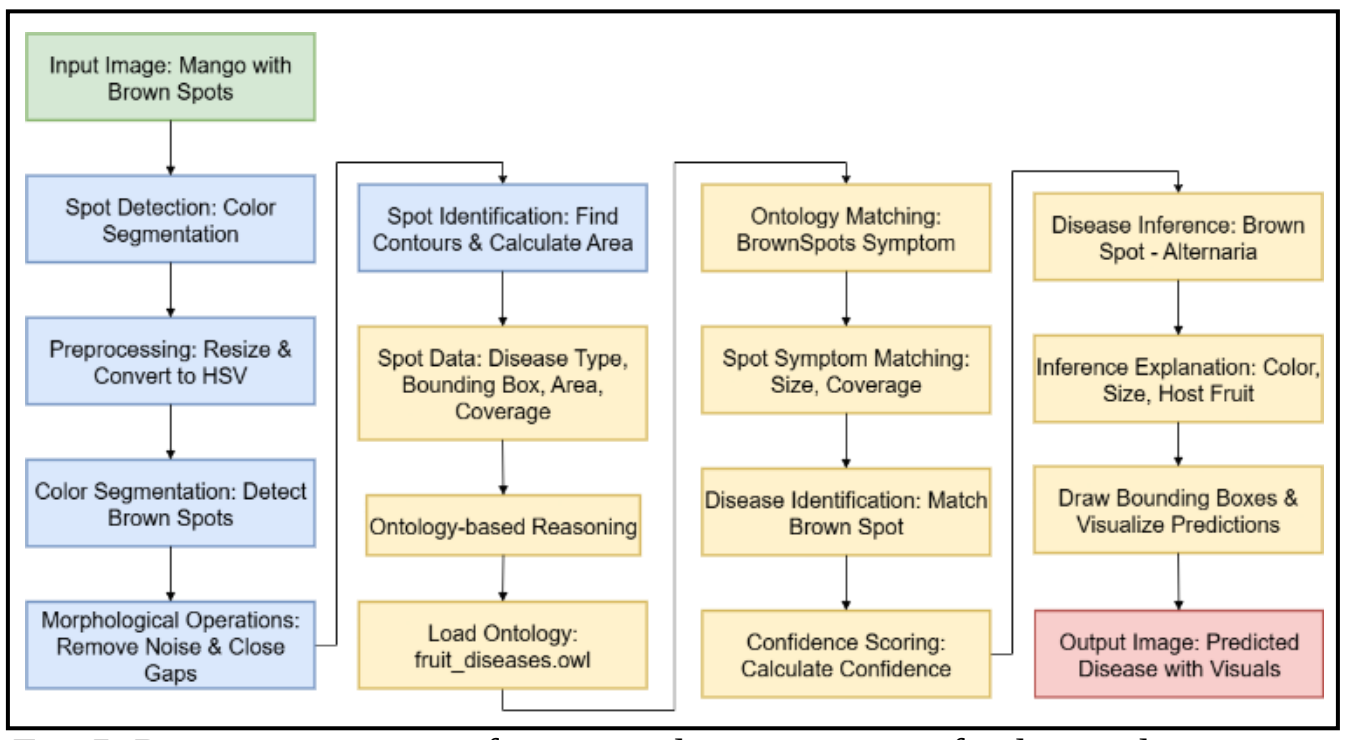

Fig. 5. Reasoning process of an example image input of a diseased mango.

The workflow diagram presents a clear flow layout that shows a design that follows a sequential path with an introduction of an input image and continues to the production of diagnosis outputs. The process is activated by a mango that has brown spots. An early stage uses a vision pipeline, the image is pre-processed using resize and transformation to the HSV color space, which is known to withstand illumination variations better. Color segmentation then isolates possible disease areas and morphological operations are used to refine these sections establishing noise and filling gaps.

As soon as the spots are outlined, the system derives the quantitative characteristics, such as area, bounded box, and percentage of the coverage. This information is subsequently relayed to ontology driven reasoning engine that is loaded with the fruit_diseases.owl knowledge base. The engine then finds general visual evidence that matches the general symptom such as BrownSpots and then it compares the retrieved feature values to a set of parameters that have been defined relating to such diseases that cause such a symptom. Based on such a match, a confidence score is computed; other important grounds include match certainty, host-pathogen specificity (establishing the disease in the suspected fruit). Lastly, the engine reports the disease (e.g., Alternaria), it calculates the rationale behind its decision and returns the original image marked with bounding boxes, predicted disease labels and their confidence levels.

## IV. Results and Evaluation

### A. *Ontology Quality Assessment*

In order to conduct a systematic assessment of the quality, stability, and utility of the designed fruit disease ontology, a multidimensional computational algorithm was used. This framework, which has been applied to Python, allows a structured evaluation of the ontology in four dimensions namely syntactic, semantic, structural and pragmatic, thus providing a comprehensive evaluation of quantitative and qualitative analysis. With the help of this framework, objective and replicable measurements were received and the subjective inspection was essentially replaced.

TABLE I. EVALUATION RESULTS OF THE DEVELOPED SYSTEM

| Dimension | Score (/10) | Interpretation |
|---|---|---|
| **Syntactic** | 8.0 | Well documented and consistent concept naming. |
| **Semantic** | 7.2 | Good conceptual richness in terms of relationships and attributes. |
| **Structural** | 7.3 | Logically sound and well-organized class property architecture. |
| **Pragmatic** | 7.0 | Adequate for domain specific computer vision applications. |
| **Overall Quality** | **7.4** | **Strong, detailed, and practically useful for automated fruit disease identification.** |

The thorough use of the fruit disease ontology resulted in a complex of the quantitative indicators, where the overall result is equal to 7.4 out of 10. In all the four dimensions, the ontology rated at the higher spectrum of the associated scales namely syntactic dimension, semantic dimension, structural dimension, and pragmatic dimension. The pragmatic ranking used to determine how appropriate the ontology is to use in computer vision applications gave the result 7.0 which means that the ontology is sufficient in describing some of the domain concepts and gives the required detection algorithms data properties. Taken together, these good scores of quality indicate that the fruit disease ontology is rich, comprehensive, and indeed useful in the identification of the automated identification of fruit diseases.

### B. *Overall System Performance Benchmarking*

It is not only the ontological representation that determines the accuracy of any end to end diagnostic system but also the ability to have high predictive performance. To test this ability in the real world setting, the system was assessed in the MangoFruitDDS dataset [11] available publicly on Mendeley where there were visual data samples used that represented a range of mango diseases. The comparison between the results and three well-defined deep learning benchmarks, namely ResNet50, VGG16 and a Multi-Layer Perceptron (MLP) baseline, was conducted in terms of the standard classification metrics namely Precision, Recall, and F1 Score.

As indicated in Table 2 that shows the comparative findings, the ontology-driven system produces results that match or exceed the ones produced by each of the comparative models. In particular, the system will provide Precision equal to 85.00%, Recall equal to 85.23% and F1 Score equal to 85.00%.

TABLE II. EVALUATION RESULTS OF THE DEVELOPED SYSTEM

| Model | Precision (%) | Recall (%) | F1 Score (%) |
|---|---|---|---|
| ResNet50 | 89.09 | 90.79 | 89.82 |
| VGG16 | 85.34 | 85.89 | 85.48 |
| **Ontology Based Vision System** | **85.00** | **85.23** | **85.00** |
| MLP (baseline) | 66.32 | 66.67 | 66.03 |

Even though ResNet50 obtains the best scores, the ontology-based approach shows a rather competitive performance that is significantly comparable to VGG16. In addition to this, it is also superior to the baseline MLP, hammering on the fact that it is really essential to incorporate a formal and semantically marked ontology into diagnostic reasoning. Those results suggest that, by being a hybrid architecture, the system is comparable in terms of diagnostic accuracy, yet it is also transparent and adaptable, which are some of the crucial properties in purely data-driven models that are typically desirable.

## V. DISCUSSION AND CONCLUSION

The paper proposes a hybrid system that can effectively combine the formal ontology and a computer vision into a transparent fruit disease detection system that can serve as a bright substitute to the opaque so called black box models. The process of diagnostic is generated based on verifiable rules and the quality of the knowledge base used is proven by the fact that the computational score is 7.4/10. Evaluation of performance on the MangoFruitDDS dataset [11] also underlines the feasibility of this methodology, with an F1

Score equal to 85.00%, extremely competitive on the market with existing deep learning models like VGG16. It is the main strength of using this type of approach, the benefit of which goes beyond the dexterity of attaining high accuracy to the benefit of offering a more solid, explainable, and flexible solution that need not rely on the immense, labeled datasets to retrain itself and hence be much more practical and less costly to manage in the given aspect of agriculture.

To sum up, this study effectively proves that the ontology-driven framework can fill this severe gap between diagnostic accuracy and transparency to provide the foundation of the next generation of credible intelligent agricultural instruments. Our future research plans include continued development of the system through adding multi modal inputs including texture and environmental information to the ontology, integrating probabilistic models of reasoning into the system to improve the ability to reason in the face of uncertainty, as well as an augmented comparative analysis over a larger and more varied dataset to further justify the system as being generalizable.